\documentclass[10pt,twocolumn,letterpaper]{article}
\usepackage{svg}
\usepackage{caption} 
\usepackage{cvm}
\usepackage{times}
\usepackage{epsfig}
\usepackage{graphicx}
\usepackage{amsmath}
\usepackage{amssymb}
\usepackage{booktabs}
\usepackage{enumitem}
\usepackage[utf8]{inputenc} 
\usepackage{newunicodechar}
\usepackage{capt-of} 
\usepackage{adjustbox}
\usepackage{pifont}
\usepackage{wasysym}
\usepackage{newunicodechar}
\newunicodechar{✓}{\checkmark}
\newunicodechar{✗}{\ding{55}}
\newunicodechar{○}{\Circle}
\usepackage{multirow}
\usepackage{tabularx}
\usepackage{rotating}   
\usepackage{pdflscape}
\usepackage{subcaption}
\usepackage{float} 
\usepackage{placeins}
\usepackage{cuted}

\usepackage[pagebackref=true,breaklinks=true,letterpaper=true,colorlinks,bookmarks=false]{hyperref}

\newcommand{\keywords}[1]{{\bf \emph{Keywords: #1}}}

\cvmfinalcopy

\def\cvmPaperID{****} 

\ifcvmfinal\fi
\begin{document}

\title{From Far and Near: Perceptual Evaluation of Crowd Representations Across Levels of Detail}

\author{
Xiaohan Sun\\
Trinity College Dublin\\
Ireland\\
{\tt\small sunx4@tcd.ie}
\and
Carol O'Sullivan\\
Trinity College Dublin\\
Ireland\\
{\tt\small Carol.OSullivan@tcd.ie}
}

\twocolumn[{
\maketitle
\begingroup
\centering
\includegraphics[width=\textwidth]{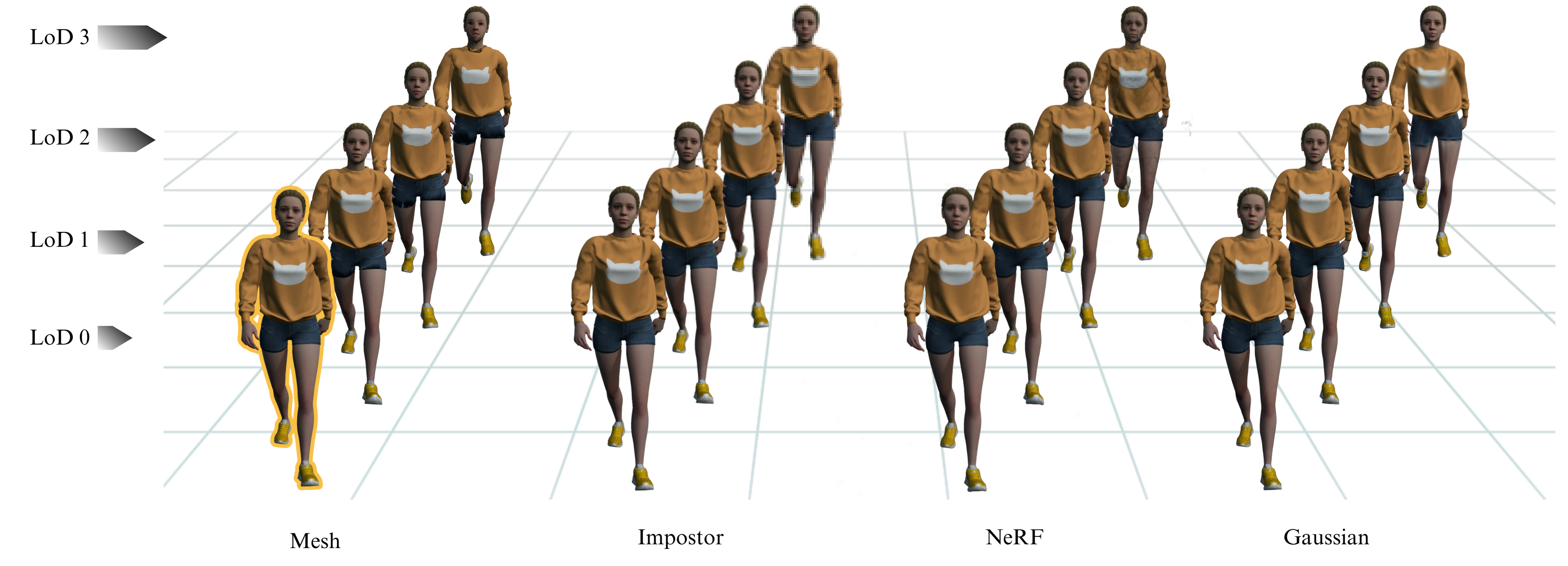}
\captionsetup{aboveskip=0.5em, belowskip=0.75em}
\captionof{figure}{\small Comparison of four crowd rendering representations (geometric meshes, image-based impostors, Neural Radiance Fields, and 3D Gaussians) across four levels of detail (LoD 0-3). LoD 0 corresponds to the highest resolution (100 \%), with detail halved at each subsequent level (50 \%, 25 \%, 12.5 \%). The examples illustrate how fidelity, structural clarity, and smoothness degrade under simplification, revealing characteristic differences among representation types.}
\label{fig:teaser}
\endgroup
\vspace{2.25em}
}]

\begin{abstract}
    In this paper, we investigate how users perceive the visual quality of crowd character representations at different levels of detail (LoD) and viewing distances. Each representation, including geometric meshes, image-based impostors, Neural Radiance Fields (NeRFs), and 3D Gaussians, exhibits distinct trade-offs between visual fidelity and computational performance. Our qualitative and quantitative results provide insights to guide the design of perceptually optimized LoD strategies for crowd rendering.
\end{abstract}


\section{Introduction}
Level of detail (LoD) techniques are widely used in crowd rendering to balance visual realism with performance efficiency. Simplified geometric meshes and image-based impostors have previously been employed to render large, animated crowds in real time (e.g., \cite{McDonnell2005Comparative}). With the recent emergence of neural rendering methods, such as Neural Radiance Fields (NeRF) and 3D Gaussian Splatting (3DGS), a new question is raised: how do traditional LoD approaches compare with modern neural representations; and under what conditions might a neural representation be perceptually equivalent to a high quality mesh representation?

To answer this question, we compare the perceptual impact of LoD rendering for both traditional and neural crowd character representations. Through a  controlled user study, we examine which LoD representations are perceptually indistinguishable from a high-resolution, ground-truth mesh at different viewing distances and levels of detail. Our contributions include:

\begin{enumerate}[leftmargin=*]
\item \textbf{Perceptual thresholds for LoD placement:}
We analyze the proportion of times each of the four representations—\emph{Mesh}, \emph{Impostor}, \emph{NeRF}, and \emph{3D Gaussian}—is judged as most similar to a high-quality mesh. Each representation is evaluated at multiple LoDs and viewing distances, both with and without motion. The resulting perceptual data can inform threshold selection for LoD placement and switching.

\item \textbf{Perception-driven LoD pipeline and toolkit.}
We introduce a representation-aware LoD rendering pipeline that integrates the four crowd character representations.  For reproducibility, we also provide practical guidelines and tools for generating customized LoD assets.
\end{enumerate}

\vspace{4pt}
\section{Related Work}

\subsection{Data Representations for Crowd Rendering}
Real-time crowd systems have traditionally relied on rigged, skinned \emph{meshes}. In production, assets often share a common rig and topology with per-agent variation in materials and animation clips; engines pack textures into atlases, reuse animation via clip sampling and blending, and issue large instanced draw calls with palette skinning to keep submission overhead low at scale \cite{Beacco2016Survey,Ryder2005Survey}.

\emph{Image-based} approaches replace geometry with 2D view samples, thereby decoupling per-agent cost from polygon count. Early city-scale systems involved the precomputation of atlases over azimuth and animation phases \cite{Tecchia2002ImageCrowd}. Later,  silhouette errors and texture distortions were reduced by using 2D polygonal proxies \cite{Kavan2008Polypostors}, or by animating limb imposters independently \cite{Beacco2012PerJointImpostors}. A recent neural rendering approach compresses appearance and motion with a CNN to enable constant-time, view-flexible character rendering with explicit material/lighting control \cite{Ostrek2024NeuropostorsICPR}.

\emph{Point-based} rendering methods trade connected mesh polygons for point and particle primitives, and can deliver simple, prefilterable samples that are suitable for animation. Early works established the use of points as rendering primitives \cite{Levoy1985Points} and introduced surface splatting with anisotropic filters \cite{Zwicker2001SurfaceSplatting}. For animated content, Wand et al. \cite{Wand2002MRRncludes} achieved real-time rendering of complex, moving geometry by building multi-resolution hierarchies of prefiltered point sets for keyframed scenes. Ellipsoidal splat formulations extend this idea by using anisotropic kernels with visibility-aware rasterization \cite{Zwicker2001SurfaceSplatting}. This precursor to the ellipsoidal kernels used in modern Gaussian splatting \cite{Kerbl2023TOG} is conceptually akin to point-based rendering, but with learned or optimized appearance parameters.

\emph{Neural} scene representations provide promising new LoD options beyond polygons and image-based imposters. For example, Neural Radiance Fields (NeRF) can model volumetric radiance via a per-ray multilayer perceptron (MLP) \cite{Mildenhall2020NeRF}. These are  typically converted to explicit or factorized structures for speedy deployment \cite{Yu2021PlenOctrees,mueller2022instant}. Gaussian splatting can also deliver real-time rendering using anisotropic splats and a fast, visibility-aware renderer \cite{Kerbl2023TOG}, while space–time extensions tackle dynamic scenes and temporal coherence \cite{Wu2024CVPR}.

\subsection{Level of Detail (LoD) for Animated Crowds}
LoD rendering for animated crowds may be framed as the process of choosing the least costly representation per character that is perceptually adequate for given viewing conditions. LoD strategies that consider geometry, motion, materials, and behavior have been proposed, incorporating hierarchical reductions of skeletons, meshes, and animation samples to deliver interactive rates at scale. These approaches exploit batching and instancing, GPU-centric skinning and shading, animation compression, and visibility/overdraw control \cite{OSullivan2002LoD,Toledo2014HLoD,Beacco2016Survey,Ryder2005Survey}. 

At larger viewing distances, image-based impostors can be displayed based on  the camera viewpoint and each character's animation pose, thus reducing the overall rendering cost \cite{Tecchia2002ImageCrowd}. The viable viewing distance can be further extended by preserving silhouettes using 2D polygonal proxies \cite{Kavan2008Polypostors}, while per-joint impostors provide more flexibility for varied animations \cite{Beacco2012PerJointImpostors}. Point/particle proxies (and displaced-subdivision variants) have also been proposed and compared with image-based methods \cite{Rudomin2004PointDispSubdiv}.
Hybrid 2.5D systems, such as \emph{Geopostors} \cite{Dobbyn2005Geopostors}, strive to balance memory, draw calls, and temporal plausibility .  Viewing distance bands are defined, beyond which geometry is replaced by impostors, and restored when the camera approaches. More recently, neural rendering methods for LoD control have been explored, e.g., by converting NeRF to explicit or factorized forms for faster evaluation, or via native, adaptive detail selection in Gaussian avatars and crowd pipelines \cite{Dongye2024LoDAvatar,Sun2025CrowdSplat}. 

Comparative studies have been conducted to examine the relative advantages of different LoD crowd character representations. McDonnell et~al.~\cite{McDonnell2005Comparative} systematically compared image-based impostors with low-resolution geometric meshes, with respect to appearance and motion fidelity. More recently, Sun et~al.~\cite{sun2025evaluating} examined the perceived quality of 3D Gaussian avatars based on motion, level of detail, and distance. In this paper, we extend this body of work by comparing the perceived quality of \emph{Mesh}, \emph{Impostor}, \emph{NeRF}, and \emph{3D Gaussian} representations of animated characters across different LoDs and viewing distances.

\section{Implementation Methods}

\subsection{Datasets}
\begin{figure}[h!]
  \centering
  \begin{adjustbox}{max width=\linewidth}
    \includegraphics{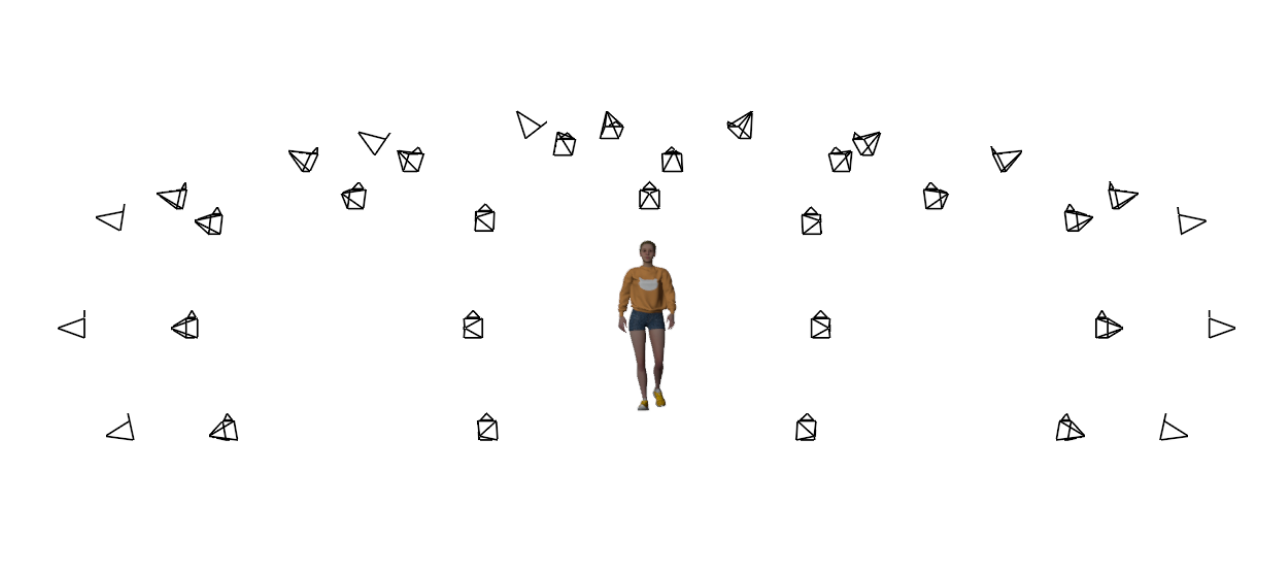}
  \end{adjustbox}
  \caption{Examples of data.}
  \label{fig:dataset}
\end{figure}

We used a Mixamo~\cite{AdobeMixamo} female character walking for 60 frames and captured each frame from 60 vitual cameras on a hemispherical rig (3{,}600 images total; see Figure~\ref{fig:dataset}). For all captures, we store standard pinhole camera intrinsics and rigid camera poses. 

To support different crowd representations, we construct representation-specific datasets from this shared capture. Mesh and impostor assets are derived directly from render-based captures, while learning-based methods additionally rely on calibrated multi-view inputs. Table~\ref{tab:data_preprocessing} summarizes the preprocessing steps applied for each representation. We next describe how this shared dataset is instantiated for each crowd representation.
\vspace{-4pt}
\begin{table}[h]
\centering
\small
\caption{Data preprocessing steps by representation used in our implementation pipeline.}
\label{tab:data_preprocessing}
\begin{tabular}{l p{0.52\linewidth}}
\toprule
\textbf{Representation} & \textbf{Data preprocessing} \\
\midrule
Mesh
& 60 cameras $\times$ 60 frames render capture; export LoDs via Blender \textit{Decimate}. \\

Impostor
& Union--alpha bounding box stabilization; pack $6\times10$ sprite sheets per LoD. \\

NeRF (Instant-NGP \cite{mueller2022instant})
& Camera intrinsics \& poses from COLMAP; graphics convention export (x right, y up, $-z$). \\

3D Gaussian (3DGS \cite{Kerbl2023TOG})
& Camera intrinsics \& poses from COLMAP; sparse colored SfM to initialize splats. \\
\bottomrule
\end{tabular}
\end{table}

\subsection{Mesh}
\begin{figure}[h]
  \centering
  \begin{adjustbox}{max width=\linewidth}
    \includegraphics{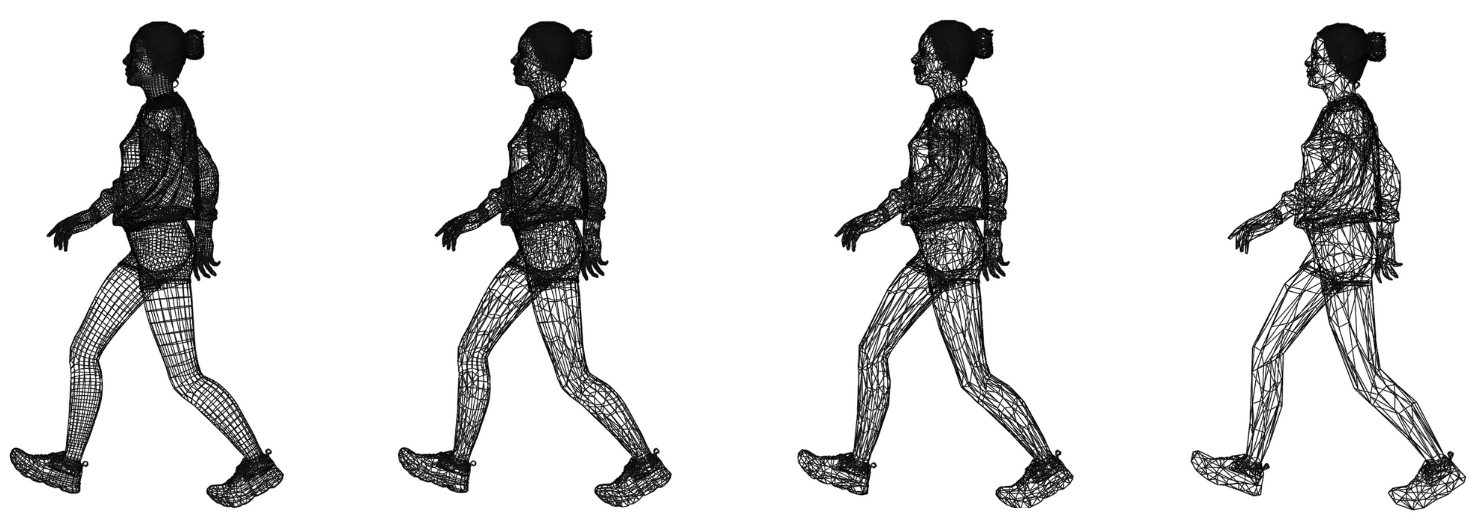}
  \end{adjustbox}
  \caption{Examples of mesh LoD.}
  \label{fig:mesh_LoD}
\end{figure}

\vspace{-4pt}
The mesh representation serves as our high-fidelity baseline, in which the captured character is rendered directly using standard skinned mesh animation. Our four Mesh LoDs are generated using Blender’s Decimate modifier. The base model (L0) and three simplified versions are created by applying decimation ratios of 1.0, 0.5, 0.25, and 0.125, yielding face counts of 27,048; 18,436; 10,811; and 5,864 respectively. Reducing the  polygon count of animated characters to generate geometric mesh LoDs is a common practice to improve rendering performance in crowds \cite{Beacco2016Survey, dong2019real}. Edge-collapse or “collapse” type decimation (as used in Blender) provides control over face count reduction while preserving shape and animation fidelity as shown in Figure~\ref{fig:mesh_LoD}.

\subsection{Impostor}
For our second representation, the animated character is rendered as a precomputed impostor sequence packed into a 6\,$\times$\,10 sprite sheet per LoD. At the highest level (L0), each tile consists of $1080{\times}1080$\ pixels, which matches the resolution used to render the reference mesh. Each lower level halves the resolution to $540{\times}540$, $270{\times}270$, and $135{\times}135$\,px (L1--L3). To avoid the per–frame ``breathing'' and root–motion loss typical of naive per–frame crops, we first compute a \emph{union} alpha bounding box over all frames of the run cycle, crop every frame to that fixed window, and apply a single global scale per LoD before centering on a square canvas. The stabilized frames are then packed column-major into a texture atlas that Blender reads as a single image. At render time we use an unlit material and animate UVs over a sprite-sheet atlas: the UVs are cropped to a single tile and offset at each frame so that the impostor displays one precomputed pose at a time. This workflow follows standard impostor practice, replacing distant geometry by a textured quad while indexing an atlas of animation frames. It is widely used in crowd rendering for performance, with quality governed by atlas resolution and sampling order \cite{Tecchia2002ImageCrowd}. To match the baked appearance, we fix the impostor’s pose and world position to the capture setup and render from the same camera transform and FOV used during capture.

\subsection{NeRF}
\vspace{2pt}
We implement NeRF as our third representation, using \emph{Instant-NGP}, which replaces the original fully connected NeRF with a multiresolution hash-grid encoding feeding a compact MLP, substantially reducing training and inference cost while retaining high visual quality \cite{Mildenhall2020NeRF,mueller2022instant}. 
LoD is controlled via the hash table capacity and network width: we fix the number of hash levels $L$ and features per level $F$, and from L0 to L3 successively halve the hash capacity (decreasing $\log_2 T$ by 1 per step) while modestly narrowing the MLP. Exact presets appear in Table~\ref{tab:LoDpresets}. 
For animation, we reconstruct independent NeRFs for each of 60 frames and render them from a shared camera path, yielding a frame-wise dynamic sequence. 
To validate that L0 serves as a suitable upper bound, we evaluated it on 60 held-out views, obtaining PSNR $=36.18$ and SSIM $=0.988$, indicating high-fidelity reconstruction quality overall.

\begin{table*}[t]
\centering
\caption{Instant-NGP LoD presets. \(L\): hash levels; \(F\): features/level; \(T = 2^{\texttt{log2\_hashmap\_size}}\). }

\label{tab:LoDpresets}
\setlength{\tabcolsep}{5pt}
\resizebox{\textwidth}{!}{%
\begin{tabular}{lcccccccc}
\toprule
\textbf{LoD} & \(\mathbf{L}\) & \(\mathbf{F}\) & \(\boldsymbol{\log_2 T}\) & \textbf{Base res.} & \textbf{Density MLP (neurons, layers)} & \textbf{Dir. enc.} & \textbf{RGB net (neurons, layers)}\\
\midrule
\textbf{LoD-0} & 12 & 2 & 18 & 16 & 128, \(1\times\) & SH deg.\,4 + Identity & 64, \(2\times\) \\
\textbf{LoD-1} & 12 & 2 & 17 & 16 & 64, \(1\times\)  & SH deg.\,3 + Identity & 32, \(2\times\)\\
\textbf{LoD-2} & 12 & 2 & 16 & 16 & 32, \(1\times\)  & SH deg.\,2 + Identity & 16, \(2\times\)\\
\textbf{LoD-3} & 12 & 2 & 15 & 16 & 16, \(1\times\)  & SH deg.\,1 + Identity & 16, \(1\times\)\\
\bottomrule
\end{tabular}%
}
\end{table*}

\subsection{Gaussian Splatting}
\emph{3D Gaussian Splatting} is our fourth representation, with a visibility-aware rasterizer and anisotropic splats \cite{Kerbl2023TOG}. Fidelity is controlled by \emph{capping the Gaussian count} $N$. For large reductions we apply a light opacity prune ($\alpha<0.01$). Our four presets are
$N{=}\{120\text{k},\,30\text{k},\,7.5\text{k},\,1.9\text{k}\}$ for L0 to L3, resp. 
All LoDs use the same optimization recipe: an $L_1{+}\text{SSIM}$ loss, spherical-harmonic color basis of degree~2, and Adam optimizer. Each continuation runs for $20$k steps. 
LoD is applied with a simple count budget and a light opacity prune. This practical setup is compatible with recent approaches that address aliasing via scale-aware filtering and reduce model size through learned pruning \cite{Yu2024MipSplatting,Zhang2024LP3DGS}.
As with NeRF, we optimize a separate 3DGS per frame for 60 frames and render from a common camera, producing the dynamic sequence. 
L0 achieves PSNR $=36.69$ and SSIM $=0.991$ on 60 held-out views, confirming that our highest-detail preset provides a strong reference for comparisons.


\section{Experimental Design}
\subsection{User Study}

\textbf{Stimuli.}
Four representations were displayed:
Image-based Impostors (\textbf{I}), Meshes (\textbf{M}), NeRFs (\textbf{N}), and 3D Gaussians (\textbf{G}).
Four levels of detail were created: (L0=100\%, L1=50\%, L2=25\%, L3=12.5\%) and five viewing distances (D0=100\% pixels, D1=80\%, D2=60\%, D3=40\%, D4=20\%).
In each trial, the four representations were displayed side by side in counterbalanced order. Participants completed two blocks: a) \emph{Video}  (with motion) and  \emph{Image} (with no motion).


\begin{figure}[h]
    \centering
    \includegraphics[width=1\linewidth]{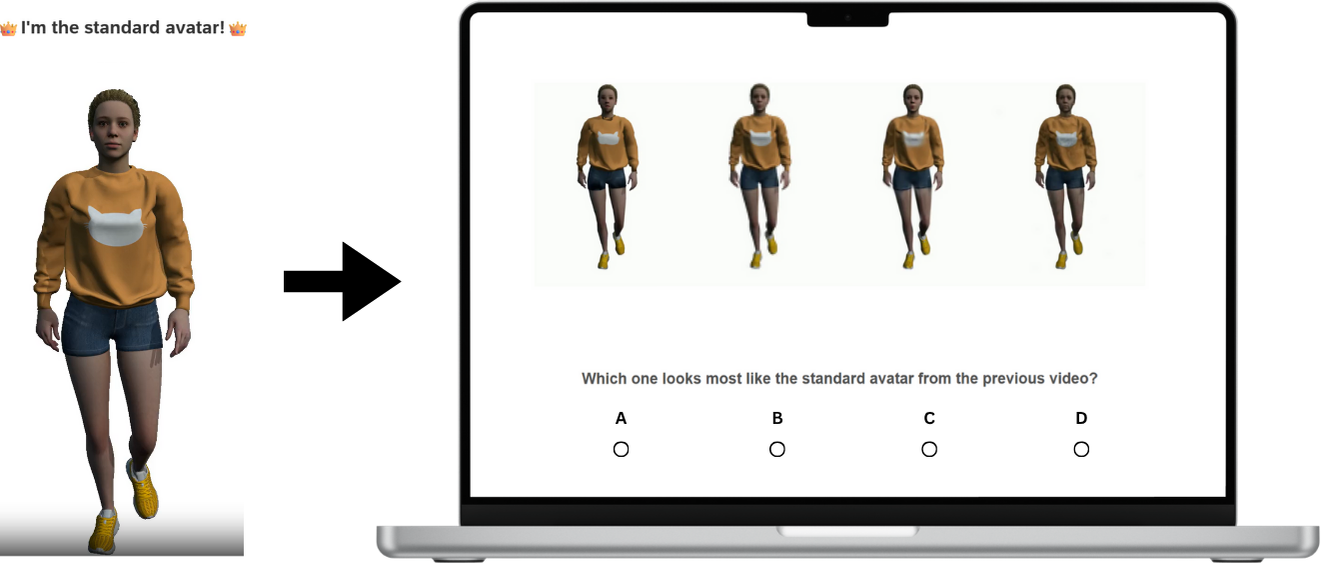}
    \caption{Example of video stimulus.}
    \label{fig:stimuli}
\end{figure}
\textbf{Participants and procedure.}
Twenty-four volunteers (17 male, 7 female; age 18–60+) completed a brief training session followed by two trial blocks (Video and Image). The experiment was conducted online via Qualtrics and was fully anonymous. Participants completed the study on their own devices (laptops, desktops, or tablets), with reported screen sizes ranging from under 15 inches to over 21 inches. Data from one participant using a mobile device were excluded to ensure consistent visual presentation. Screen size was recorded at the start of the experiment and showed no significant effect on perceptual judgments. For each trial, participants viewed a high-resolution mesh reference and then chose with one of the four displayed stimuli most closely matched the high-quality mesh (see Fig. \ref{fig:stimuli}). Trials were randomized, and the positions of the four representations were counterbalanced across participants.

\begin{figure}[t]
    \centering
    \includegraphics[width=0.5\textwidth]{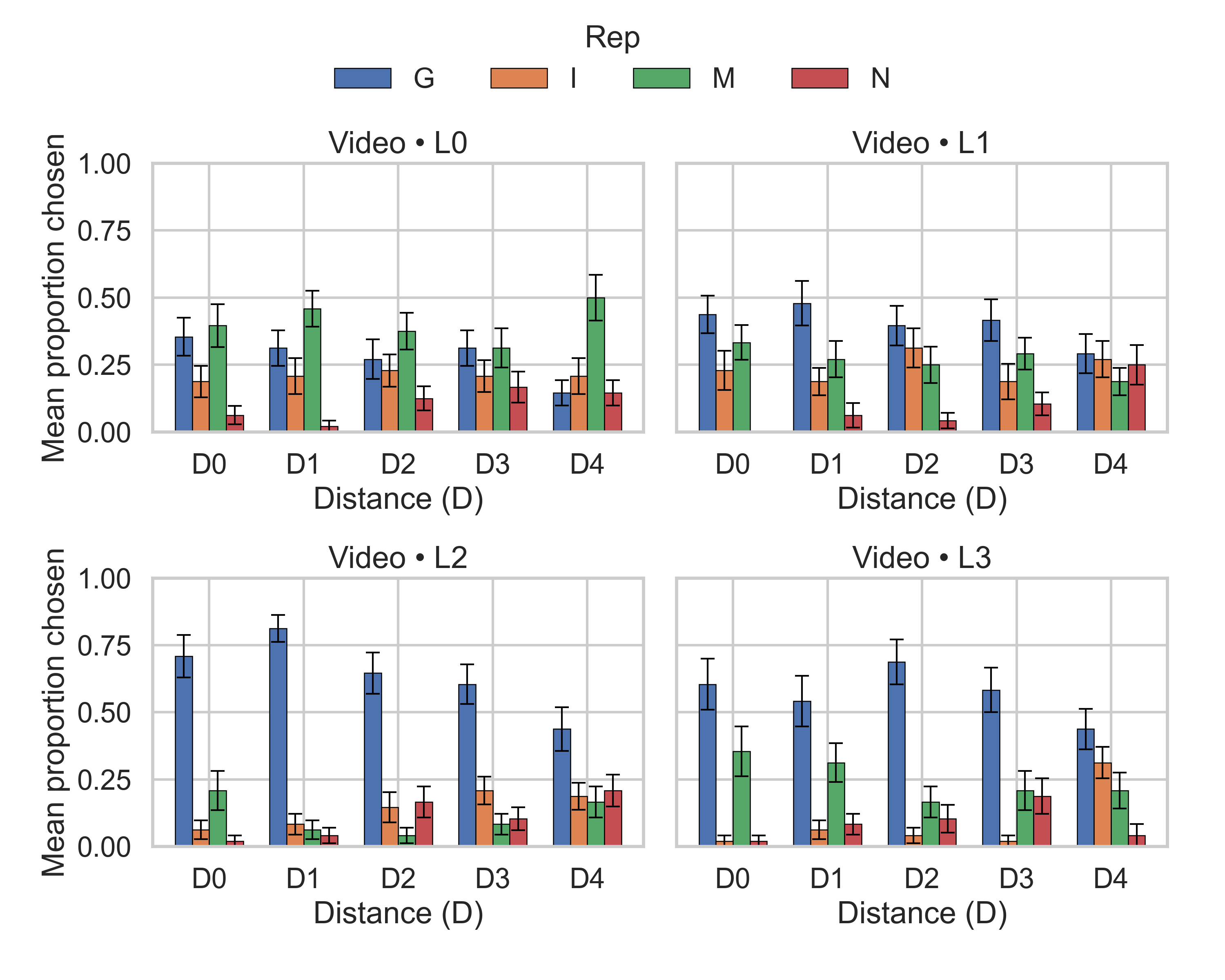}
    \caption{Video block: mean proportion chosen for each representation across LoD and Distance (D0--D4). Error bars show $\pm$\,1~SE across participants.}
    \label{fig:repr-dist-lod-video}
\end{figure}

\textbf{Statistical analysis.}
Choices were analyzed as fully within–subjects repeated measures with factors
\textit{Representation} (G, I, M, N), \textit{Distance} (D0–D4), and \textit{LoD} (L0–L3), in both \textit{Mode}s (Image vs.\ Video).
We averaged the repetitions per participant and condition to obtain selection proportions.
We fitted an OLS with subject fixed effects and the full
Representation$\times$Distance$\times$LoD$\times$Mode factorial, and report a Type~II ANOVA (Table~\ref{tab:anova_results}).
Because responses are binary at the trial level, we additionally fitted a trial–level binomial GLM (logit) with subject fixed effects, and obtained likelihood–ratio omnibus tests by comparing the full model to hierarchically reduced models (Table~\ref{tab:glm}).
Descriptive means ($\pm$\,SE) are shown in Figs.~\ref{fig:repr-dist-lod-video}.

\textbf{Results.}
The ANOVA revealed a strong main effect of \textit{Representation}, 
as well as robust \textit{Representation}$\times$\textit{Distance} and \textit{Representation}$\times$\textit{LoD} interactions,
The three–way interaction \textit{Representation}$\times$\textit{Distance}$\times$\textit{LoD} was also significant.
No main effects of \textit{Distance}, \textit{LoD} or \textit{Mode} (Image vs.\ Video) were observed 
and no interactions involving \textit{Mode} reached significance (min $p=0.257$).
The confirmatory trial–level GLM echoed these patterns with significant omnibus LR tests for \textit{Representation}, \textit{Distance}, \textit{LoD}, \textit{Representation}$\times$\textit{LoD}, \textit{LoD}$\times$\textit{Distance}, and the three–way interaction (Table~\ref{tab:glm}).

\vspace{4pt}
\begin{table*}[t]
\centering
\caption{ANOVA showing the results of main and interaction effect tests with degrees of freedom, effect sizes ($\eta^{2}$) and ($p$) values.  (Common denominator df for the OLS Type~II ANOVA is $3657$.)}
\label{tab:anova_results}
\small
\small
\begin{tabular}{llllll}\hline\hline
\textbf{Effect Tested} & \emph{dof} & \textbf{F-Test} & \textbf{$\eta^{2}$} & \textbf{p}\\\hline\hline
  Representation & 3 & 177.52 & 0.127 & {\color{red}$<0.001$}\\
  Representation × Distance & 12 & 8.53 & 0.027 & {\color{red}$<0.001$}\\
  Representation × LoD & 9 & 33.16 & 0.075 & {\color{red}$<0.001$}\\
  Representation × Mode & 3 & 1.08 & 0.001 & $0.356$\\
  Representation × Distance × LoD & 36 & 2.26 & 0.022 & {\color{red}$<0.001$}\\
  Representation × Distance × Mode & 12 & 0.36 & 0.001 & $0.976$\\
  Representation × LoD × Mode & 9 & 0.90 & 0.002 & $0.527$\\
  Representation × Distance × LoD × Mode & 36 & 1.14 & 0.011 & $0.257$\\
\hline\hline
\end{tabular}

\end{table*}

\begin{table}[t]
\centering
\caption{Omnibus likelihood–ratio (LR) tests from the trial–level binomial GLM with subject fixed effects.}
\label{tab:glm}
\small
\small
\begin{tabular}{lrrl}
\toprule
Effect & LR & df & p \\
\midrule
Representation & 1089.30 & 139 & \textcolor{red}{$< 0.001$} \\
Distance & 275.81 & 140 & \textcolor{red}{$< 0.001$} \\
LoD & 418.92 & 136 & \textcolor{red}{$< 0.001$} \\
Mode & 66.02 & 80 & $ 0.869$ \\
Representation $\times$ Distance & 61.83 & 76 & $ 0.880$ \\
Representation $\times$ LoD & 418.92 & 121 & \textcolor{red}{$< 0.001$} \\
Representation $\times$ Mode & 66.02 & 79 & $ 0.851$ \\
LoD $\times$ Distance & 152.80 & 124 & \textcolor{red}{$ 0.040$} \\
Representation $\times$ LoD $\times$ Distance & 150.97 & 112 & \textcolor{red}{$ 0.008$} \\
\bottomrule
\end{tabular}

\end{table}


\vspace{8pt}
\textbf{Discussion.}
Perceived fidelity is seen to depend on a combination of \emph{Representation}, viewing \emph{Distance}, and \emph{LoD}.
As expected, \textbf{Mesh} is most preferred at high detail and near distances, whereas \textbf{Gaussians} become increasingly indistinguishable from the Mesh as detail decreases and/or viewing distance increases.
Both the omnibus ANOVA and the confirmatory GLM reveal strong effects of \textit{Representation}, clear interactions with \textit{LoD}, and a significant three-way interaction,  whereas no significant main or interaction effect was found for \textit{Mode} (Image vs.\ Video).

\vspace{-2pt}
\subsection{Quantitative Experiments}
\vspace{-2pt}
\subsubsection{Visual Comparison}
Image fidelity is calculated using an evaluation view camera that combines poses matching the user-study setup but is excluded from NeRF/3DGS training. For each LoD and representation, we compare with the Mesh reference in the same pose using PSNR, SSIM and LPIPS \cite{wolf2009psnr, wang2004ssim, zhang2018lpips}. Table~\ref{tab:qual_metrics} reports metrics across LoDs; higher PSNR/SSIM and lower LPIPS indicate closer agreement with the Mesh baseline.

\paragraph{Summary.}
At L0, both \textbf{3D Gaussians} and \textbf{NeRF} closely match the Mesh (Gaussians: 36.69\,dB / 0.991 / 0.013; NeRF: 36.18\,dB / 0.988 / 0.019 for PSNR / SSIM / LPIPS). With decreasing LoD, \textbf{NeRF} degrades smoothly in PSNR while maintaining high SSIM and low LPIPS; \textbf{Gaussians} show a larger PSNR drop but remain competitive on SSIM/LPIPS through mid LoDs. \textbf{Impostors} exhibit high SSIM and very low LPIPS at the matched view, despite lower PSNR, consistent with view-aligned textures that preserve structure but lack 3D parallax. Overall, frame-based metrics indicate that neural fields (NeRF/3DGS) are closest to Mesh at high detail, while impostors can appear perceptually similar at single views even with lower pixel-wise fidelity.

\begin{table}[h]
\centering
\small
\caption{Image Similarity at Evaluation View.}
\label{tab:qual_metrics}
\begin{tabular}{lcccc}
\toprule
\textbf{Rep.} & \textbf{LoD} & \textbf{PSNR$\;\uparrow$} & \textbf{SSIM$\;\uparrow$} & \textbf{LPIPS$\;\downarrow$} \\
\midrule
NeRF & L0 & 36.18 & 0.988 & 0.019 \\
     & L1 & 35.08 & 0.984 & 0.022 \\
     & L2 & 33.25 & 0.977 & 0.030 \\
     & L3 & 28.09 & 0.954 & 0.052 \\
\midrule
3DGS & L0 & 36.69 & 0.991 & 0.013 \\
     & L1 & 24.76 & 0.961 & 0.029 \\
     & L2 & 24.52 & 0.954 & 0.045 \\
     & L3 & 24.01 & 0.938 & 0.090 \\
\midrule
Impostor & L0 & 23.95 & 0.965 & 0.006 \\
         & L1 & 23.40 & 0.970 & 0.010 \\
         & L2 & 21.35 & 0.970 & 0.017 \\
         & L3 & 19.40 & 0.971 & 0.030 \\
\bottomrule
\end{tabular}
\end{table}

\newpage
\begin{strip}
\centering
\small
\captionof{table}{On-disk asset size and offline per-frame training time by representation and LoD.
Training time is reported for learning-based methods only (20k steps on RTX~4090).}
\label{tab:mem_and_training}
\begin{tabular}{lcccccccc}
\toprule
\multirow{2}{*}{\textbf{Representation}} &
\multicolumn{2}{c}{\textbf{LoD0}} &
\multicolumn{2}{c}{\textbf{LoD1}} &
\multicolumn{2}{c}{\textbf{LoD2}} &
\multicolumn{2}{c}{\textbf{LoD3}} \\
\cmidrule(lr){2-3} \cmidrule(lr){4-5} \cmidrule(lr){6-7} \cmidrule(lr){8-9}
 & \textbf{Size} & \textbf{Train} &
   \textbf{Size} & \textbf{Train} &
   \textbf{Size} & \textbf{Train} &
   \textbf{Size} & \textbf{Train} \\
\midrule
NeRF (Instant-NGP)
 & 21.2\,MB & $\sim$240\,s
 & 17.0\,MB & $\sim$207\,s
 & 15.5\,MB & $\sim$214\,s
 & 14.0\,MB & $\sim$205\,s \\

3D Gaussian (3DGS)
 & 19.0\,MB & $\sim$320\,s
 & 8.00\,MB & $\sim$122\,s
 & 2.00\,MB & $\sim$78\,s
 & 0.768\,MB & $\sim$58\,s \\

Impostor
 & 328\,KB & ---
 & 100\,KB & ---
 & 32.0\,KB & ---
 & 12.0\,KB & --- \\

Mesh
 & 5.19\,MB & ---
 & 4.37\,MB & ---
 & 3.86\,MB & ---
 & 3.50\,MB & --- \\
\bottomrule
\end{tabular}
\end{strip}

{\subsubsection{Memory Usage and Training Time}}
We quantify deployment memory as the on-disk size of the exact LoD assets used in the study. This metric is engine-agnostic and comparable across heterogeneous toolchains.

Table~\ref{tab:mem_and_training} reports the footprint per representation and LoD. The entries correspond to (i) \textbf{Mesh}: the rigged mesh package (geometry, armature, and textures), (ii) \textbf{Impostor}: billboard atlas images, (iii) \textbf{3D Gaussian}: optimized Gaussian caches, and (iv) \textbf{NeRF}: trained model snapshots.

For learning-based representations, we additionally report the offline training time required to generate each LoD asset. 
This cost reflects authoring and preprocessing overhead and is incurred prior to deployment, with no impact on runtime performance. 
Training time is reported per frame. 
Traditional representations (meshes and impostors) do not require learning-based optimization and therefore have no associated training cost.

\paragraph{Summary.}
Deployment footprint varies significantly by representation. \textbf{Impostors} are the most memory-efficient by a wide margin and scale down aggressively with decreasing LoD, while also incurring no learning-based training cost. \textbf{3D Gaussian} representations exhibit strong compressibility across LoDs, offering a tunable volumetric alternative that requires substantially less memory than meshes at lower LoDs, at the expense of offline training time that decreases with representation complexity. In contrast, \textbf{Mesh} assets yield only modest memory savings under simplification, reflecting constraints imposed by geometry and rig data, but benefit from the absence of training overhead. \textbf{NeRFs} require the largest deployment footprint across all LoDs and achieve comparatively smaller proportional memory reductions, while also incurring consistent offline training costs per LoD. Overall, these results highlight distinct trade-offs between memory efficiency and authoring cost across representations, which are critical to consider when designing perceptually optimized LoD strategies for animated crowds.

\section{Conclusions and Future Work}
\vspace{10pt}
Our results show that perceived similarity to a mesh ground truth is driven primarily by a character's \emph{Representation} and its interactions with \emph{Distance} and \emph{LoD}, and does not appear to be affected by the presence or absence of motion. 

Based on these results, we propose the following practical guidelines: 
\textbf{Impostors} are the cheapest and most scalable choice for far/low-detail rendering (i.e., minimal geometry cost and excellent batching at distance), although flexibility is sacrificed for large precomputed atlases and limited animation blending. \textbf{3D Gaussians} become increasingly indistinguishable from meshes as pixel density drops or detail is reduced, thus offering a strong mid-to-high LoD alternative with real-time rendering.  Geometric \textbf{Meshes} retain a clear advantage at nearer views and higher detail, but for further distances, more efficient representations are highly competitive. 

Our findings have direct implications for immersive and virtual reality applications, where perceptually informed LoD selection can help balance visual realism and performance in crowd-populated environments, contributing to stable frame rates and user comfort. We conclude by outlining directions for future work.

\textbf{Beyond appearance: motion, dynamics, and generalization.}
To enable a controlled comparison, our study focuses on visual appearance using a single character, fixed lighting, and a simple animation. Within this setting, motion does not significantly affect perceived similarity across representations. An interesting direction for future work is to extend this evaluation to richer motion patterns, diverse character appearances and lighting conditions, and physical interactions such as clothing dynamics and inter-agent contact. Longer sequences and more dynamic scenarios may further reveal how perceptual trends observed here carry over to increasingly complex crowd settings, and motivate the exploration of temporally consistent and dynamic neural rendering approaches.

\newpage
\textbf{Runtime performance and system-level evaluation.}
While this work emphasizes perceptual quality and asset-level efficiency, end-to-end runtime performance remains a critical factor for real-time deployment and depends on hardware, renderer implementation, and system-level optimizations. Future work should incorporate standardized performance benchmarks to more explicitly quantify the trade-offs between perceptual fidelity and rendering throughput in practical deployment scenarios.

In summary, we present a unified perceptual evaluation of Mesh, Impostor, NeRF, and 3D Gaussian representations and provide practical guidance for managing LoD in crowd rendering systems. Our results and accompanying pipeline offer a foundation for designing perceptually driven crowd representations that scale effectively across viewing conditions.



\section*{Acknowledgement}
This work was conducted with the financial support of the Research Ireland Centre for Research Training in Digitally-Enhanced Reality (d-real) under Grant No. 18/CRT/6224. For the purpose of Open Access, the author has applied a CC BY public copyright licence to any Author Accepted Manuscript version arising from this submission.

{\small
\bibliographystyle{cvm}
\bibliography{cvmbib}

@article{mueller2022instant,
  title={Instant Neural Graphics Primitives with a Multiresolution Hash Encoding},
  author={M{\"u}ller, Thomas and Evans, Alex and Schied, Christoph and Keller, Alexander},
  journal={ACM Transactions on Graphics (SIGGRAPH)},
  volume={41}, number={4}, pages={1--15},
  year={2022},
  doi={10.1145/3528223.3530127},
  url={https://arxiv.org/abs/2201.05989}
}

@article{dong2019real,
  title={Real-Time Large Crowd Rendering with Efficient Character and Instance Management on GPU},
  author={Dong, Yangzi and Peng, Chao},
  journal={International Journal of Computer Games Technology},
  volume={2019},
  number={1},
  pages={1792304},
  year={2019},
  publisher={Wiley Online Library}
}

@article{Beacco2016Survey,
  author    = {A. Beacco and N. Pelechano and C. And{\'u}jar},
  title     = {A Survey of Real-Time Crowd Rendering},
  journal   = {Computer Graphics Forum},
  year      = {2016},
  volume    = {35},
  number    = {8},
  pages     = {32--50},
  doi       = {10.1111/cgf.12774}
}

@incollection{Ryder2005Survey,
  author    = {G. Ryder and M. Chalmers and R. McCready},
  title     = {A Survey of Real-Time Rendering Techniques for Crowds},
  booktitle = {Eurographics State of the Art Reports},
  year      = {2005}
}

@inproceedings{McDonnell2005Comparative,
  author    = {Rachel McDonnell and Simon Dobbyn and Carol O'Sullivan},
  title     = {LOD Human Representations: A Comparative Study},
  booktitle = {V-CROWDS: Intl. Workshop on Crowd Simulation},
  year      = {2005},
  pages     = {101--115},
  url       = {https://www.scss.tcd.ie/rachel.mcdonnell/papers/VCrowds.pdf}
}

@article{OSullivan2002LOD,
  author    = {Carol O'Sullivan and Justine Cassell and Jana Dingliana and Simon Dobbyn and Barry McNamee and Chris Peters and Thinh Giang},
  title     = {Levels of Detail for Crowds and Groups},
  journal   = {Computer Graphics Forum},
  year      = {2002},
  volume    = {21},
  number    = {4},
  pages     = {733--741},
  doi       = {10.1111/1467-8659.00631}
}

@article{Toledo2014HLOD,
  author    = {Leonel Toledo and Oscar De Gyves and Victor Rudom{\'i}n},
  title     = {Hierarchical level of detail for varied animated crowds},
  journal   = {The Visual Computer},
  year      = {2014},
  volume    = {30},
  number    = {6},
  pages     = {949--961},
  doi       = {10.1007/s00371-014-0975-9}
}

@inproceedings{Mildenhall2020NeRF,
  author    = {Ben Mildenhall and Pratul P. Srinivasan and Matthew Tancik and Jonathan T. Barron and Ravi Ramamoorthi and Ren Ng},
  title     = {NeRF: Representing Scenes as Neural Radiance Fields for View Synthesis},
  booktitle = {ECCV},
  year      = {2020},
  url       = {https://arxiv.org/abs/2003.08934}
}

@article{Kerbl2023TOG,
  author    = {Bernhard Kerbl and Georgios Kopanas and Thomas Leimk{\"u}hler and George Drettakis},
  title     = {3D Gaussian Splatting for Real-Time Radiance Field Rendering},
  journal   = {ACM Transactions on Graphics},
  year      = {2023},
  volume    = {42},
  number    = {4},
  pages     = {1--14},
  doi       = {10.1145/3592433},
  url       = {https://repo-sam.inria.fr/fungraph/3d-gaussian-splatting/}
}

@misc{AdobeMixamo,
  author       = {{Adobe}},
  title        = {Mixamo},
  howpublished = {\url{https://www.mixamo.com/}},
  note         = {Accessed 2025-10-13}
}

@inproceedings{Wu2024CVPR,
  author    = {Guanjun Wu and Taoran Yi and Jiemin Fang and Lingxi Xie and Xiaopeng Zhang and Wei Wei and Wenyu Liu and Qi Tian and Xinggang Wang},
  title     = {4D Gaussian Splatting for Real-Time Dynamic Scene Rendering},
  booktitle = {CVPR},
  year      = {2024},
  pages     = {},
  url       = {https://openaccess.thecvf.com/content/CVPR2024/papers/Wu_4D_Gaussian_Splatting_for_Real-Time_Dynamic_Scene_Rendering_CVPR_2024_paper.pdf}
}

@inproceedings{Yu2021PlenOctrees,
  author    = {Yu, Alex and Li, Ruilong and Tancik, Matthew and Hao, Li and Ng, Ren and Kanazawa, Angjoo},
  title     = {PlenOctrees for Real-Time Rendering of Neural Radiance Fields},
  booktitle = {ICCV},
  year      = {2021}
}

@article{Tecchia2002ImageCrowd,
  author  = {Tecchia, Franco and Loscos, C{\'e}line and Chrysanthou, Yiorgos},
  title   = {Image-Based Crowd Rendering},
  journal = {IEEE Computer Graphics and Applications},
  year    = {2002},
  volume  = {22},
  number  = {2},
  pages   = {36--43},
  doi     = {10.1109/38.988745}
}

@inproceedings{Kavan2008Polypostors,
  author    = {Kavan, Ladislav and Dobbyn, Simon and Collins, Steven and {\v{Z}}{\'a}ra, Ji{\v{r}}{\'\i} and O'Sullivan, Carol},
  title     = {Polypostors: 2D Polygonal Impostors for 3D Crowds},
  booktitle = {Proceedings of the Symposium on Interactive 3D Graphics and Games (I3D)},
  year      = {2008},
  pages     = {149--155},
  doi       = {10.1145/1342250.1342273}
}

@article{Beacco2012PerJointImpostors,
  author  = {Beacco, {\'A}lvaro and Pelechano, N{\'u}ria and And{\'u}jar, Carlos},
  title   = {Efficient rendering of animated characters through optimized per-joint impostors},
  journal = {Computer Animation and Virtual Worlds},
  year    = {2012},
  volume  = {23},
  number  = {3--4},
  pages   = {331--339},
  doi     = {10.1002/cav.1422}
}

@inproceedings{Rudomin2004PointDispSubdiv,
  author    = {Rudom{\'\i}n, Isaac and Mill{\'a}n, Erik},
  title     = {Point-based Rendering and Displaced Subdivision for Interactive Animation of Crowds of Clothed Characters},
  booktitle = {VRIPHYS 2004: Virtual Reality Interaction and Physical Simulation Workshop},
  year      = {2004},
  pages     = {139--148}
}

@inproceedings{Dobbyn2005Geopostors,
  author    = {Dobbyn, Simon and Hamill, John and O'Conor, Kieran and O'Sullivan, Carol},
  title     = {Geopostors: A Real-Time Geometry/Impostor Crowd Rendering System},
  booktitle = {Proceedings of the Symposium on Interactive 3D Graphics and Games (I3D)},
  year      = {2005},
  pages     = {95--102},
  doi       = {10.1145/1053427.1053443}
}

@misc{Dongye2024LoDAvatar,
  author       = {Dongye, Xiaonuo and Guo, Hanzhi and Luo, Le and Jiang, Haiyan and Bao, Yihua and Tian, Zeyu and Weng, Dongdong},
  title        = {LoDAvatar: Hierarchical Embedding and Adaptive Levels of Detail with Gaussian Splatting for Enhanced Human Avatars},
  howpublished = {arXiv:2410.20789},
  year         = {2024},
  url          = {https://arxiv.org/abs/2410.20789}
}

@misc{Sun2025CrowdSplat,
  author       = {Sun, Xiaohan and Xu, Yinghan and Dingliana, John and O'Sullivan, Carol},
  title        = {CrowdSplat: Exploring Gaussian Splatting for Crowd Rendering},
  howpublished = {arXiv:2501.17792},
  year         = {2025},
  url          = {https://arxiv.org/abs/2501.17792},
  note         = {Includes LoD integration for crowd-scale rendering}
}

@inproceedings{yu2024mipsplatting,
  title={Mip-Splatting: Alias-free 3D Gaussian Splatting},
  author={Yu, Zehao and Kerbl, Bernhard and Xie, Chuan and R{\"u}egg, Nils and Liu, Sida I. and Geiger, Andreas},
  booktitle={CVPR},
  year={2024},
  url={https://openaccess.thecvf.com/content/CVPR2024/papers/Yu_Mip-Splatting_Alias-free_3D_Gaussian_Splatting_CVPR_2024_paper.pdf}
}

@inproceedings{zhang2024lp3dgs,
  title={LP-3DGS: Learning to Prune 3D Gaussian Splatting},
  author={Zhang, Zhaoliang and Song, Tianchen and Lee, Yongjae and Yang, Li and Peng, Cheng and Chellappa, Rama and Fan, Deliang},
  booktitle={NeurIPS},
  year={2024},
  url={https://proceedings.neurips.cc/paper_files/paper/2024/file/dd51dbce305433cd60910dc5b0147be4-Paper-Conference.pdf}
}

@techreport{wolf2009psnr,
  title        = {Reference Algorithm for Computing Peak Signal-to-Noise Ratio (PSNR) of a Video Sequence with a Constant Delay},
  author       = {Wolf, Stephen and Pinson, Margaret},
  institution  = {ITU-T Study Group 9 (contribution from NTIA/ITS, USA)},
  type         = {Contribution COM 9--C 6},
  address      = {Geneva, Switzerland},
  month        = feb,
  year         = {2009},
  url          = {https://its.ntia.gov/media/27512/itut_com9_c6.pdf},
  note         = {Defines a standardized PSNR computation procedure for video}
}

@article{wang2004ssim,
  title   = {Image Quality Assessment: From Error Visibility to Structural Similarity},
  author  = {Wang, Zhou and Bovik, Alan C. and Sheikh, Hamid R. and Simoncelli, Eero P.},
  journal = {IEEE Transactions on Image Processing},
  volume  = {13},
  number  = {4},
  pages   = {600--612},
  year    = {2004},
  doi     = {10.1109/TIP.2003.819861}
}

@inproceedings{zhang2018lpips,
  title     = {The Unreasonable Effectiveness of Deep Features as a Perceptual Metric},
  author    = {Zhang, Richard and Isola, Phillip and Efros, Alexei A. and Shechtman, Eli and Wang, Oliver},
  booktitle = {Proceedings of the IEEE/CVF Conference on Computer Vision and Pattern Recognition (CVPR)},
  pages     = {586--595},
  year      = {2018},
  doi       = {10.1109/CVPR.2018.00068}
}

@inproceedings{sun2025evaluating,
  title={Evaluating CrowdSplat: Perceived Level of Detail for Gaussian Crowds},
  author={Sun, Xiaohan and Xu, Yinghan and Dingliana, John and O’Sullivan, Carol},
  booktitle={2025 IEEE Conference on Virtual Reality and 3D User Interfaces Abstracts and Workshops (VRW)},
  pages={720--724},
  year={2025},
  organization={IEEE}
}

@incollection{Ostrek2024NeuropostorsICPR,
  title     = {Neuropostors: Neural Geometry-Aware 3D Crowd Character Impostors},
  author    = {Ostrek, Mirela and Mitra, Niloy J. and O'Sullivan, Carol},
  booktitle = {Pattern Recognition: 27th International Conference, {ICPR} 2024, Kolkata, India, December 1--5, 2024, Proceedings, Part {XXII}},
  series    = {Lecture Notes in Computer Science},
  volume    = {15322},
  publisher = {Springer},
  year      = {2024},
  pages     = {432--448},
  doi       = {10.1007/978-3-031-78312-8_29}
}

@inproceedings{Zwicker2001SurfaceSplatting,
  author    = {Matthias Zwicker and Hanspeter Pfister and Jeroen van Baar and Markus Gross},
  title     = {Surface Splatting},
  booktitle = {Proceedings of the 28th Annual Conference on Computer Graphics and Interactive Techniques (SIGGRAPH '01)},
  pages     = {371--378},
  year      = {2001},
  publisher = {ACM},
  doi       = {10.1145/383259.383300}
}

@article{Wand2002MRRncludes,
  author  = {Michael Wand and Wolfgang Stra{\ss}er},
  title   = {Multi-Resolution Rendering of Complex Animated Scenes},
  journal = {Computer Graphics Forum},
  volume  = {21},
  number  = {3},
  pages   = {483--491},
  year    = {2002},
  note    = {Proc.\ Eurographics 2002},
  doi     = {10.1111/1467-8659.t01-1-00608}
}

@techreport{Levoy1985Points,
  author      = {Marc Levoy and Turner Whitted},
  title       = {The Use of Points as a Display Primitive},
  institution = {Department of Computer Science, University of North Carolina at Chapel Hill},
  number      = {TR 85-022},
  year        = {1985},
  url         = {https://graphics.stanford.edu/papers/points/point-tr-scanned.pdf}
}
}

\end{document}